\documentclass[10pt,twocolumn,letterpaper]{article}

\usepackage[pagenumbers]{cvpr} 

\usepackage{multirow}
\usepackage{bm}

\definecolor{cvprblue}{rgb}{0.21,0.49,0.74}
\usepackage[pagebackref,breaklinks,colorlinks,allcolors=cvprblue]{hyperref}

\def\paperID{*****} 
\def\confName{CVPR}
\def\confYear{2026}

\title{Destination-Conditioned Adaptation with Semantic-Geometric Alignment for Generative Inbetweening}

\author{
Zhiyu Xu\textsuperscript{1}\quad
Wei Shang\textsuperscript{2}\quad
Yushen Xu\textsuperscript{1}\quad
Wenwen Li\textsuperscript{3}\\
Zhiyan Xu\textsuperscript{4}\quad
Pengfei Zhu\textsuperscript{1}\quad
Dongwei Ren\textsuperscript{1}\\
\textsuperscript{1}Tianjin University\quad
\textsuperscript{2}Singapore Management University\\
\textsuperscript{3}Tianjin Medical University\quad
\textsuperscript{4}Southeast University\\
}

\begin{document}
\maketitle
\begin{abstract}
Generative inbetweening aims to synthesize temporally coherent intermediate frames between a start frame and a target frame. Although pretrained image-to-video diffusion models provide strong priors for large-motion inbetweening, they are not explicitly designed to reach a specified target frame. They may therefore generate realistic individual frames while drifting from the target or producing unstable motion under large displacements and non-rigid dynamics. We propose DSG-GI, a framework that adapts pretrained latent video diffusion models through \textbf{D}estination-Conditioned Adaptation and \textbf{S}emantic-\textbf{G}eometric Alignment. Destination-Conditioned Adaptation treats the target frame as the destination of the sequence and provides temporally progressive guidance along the generated trajectory. Semantic-Geometric Alignment aligns the token-wise spatio-temporal relational geometry of backbone features with that of a frozen video foundation encoder and applies a flow-matching-aware latent flow constraint to samples near the data endpoint of the probability path. Together, these components strengthen destination guidance and preserve coherent semantic-geometric motion structure. Extensive experiments on standard and large-motion inbetweening benchmarks demonstrate improvements in temporal stability, target-frame faithfulness, and perceptual quality. Our code will be made publicly available.
\end{abstract}


\section{Introduction}
Generative inbetweening is a fundamental task in video understanding and generation. Given a start frame and a target frame, it synthesizes intermediate frames that form a smooth, visually plausible transition. Applications include slow-motion video synthesis, frame-rate conversion, animation inbetweening, video editing, and content creation. Classical interpolation methods estimate optical flow, motion kernels, or local correspondences and synthesize missing frames by warping neighboring observations~\citep{niklaus2017video,bao2019depth,jiang2018super,niklaus2020softmax,reda2022film,zhang2023extracting,li2023amt}. These methods perform well when motion is small and approximately deterministic, but often fail under large displacements, occlusions, articulated motion, or complex scene changes, where low-level correspondence alone does not uniquely determine the intermediate content.

\begin{figure}[t]
\centering
\includegraphics[width=\columnwidth]{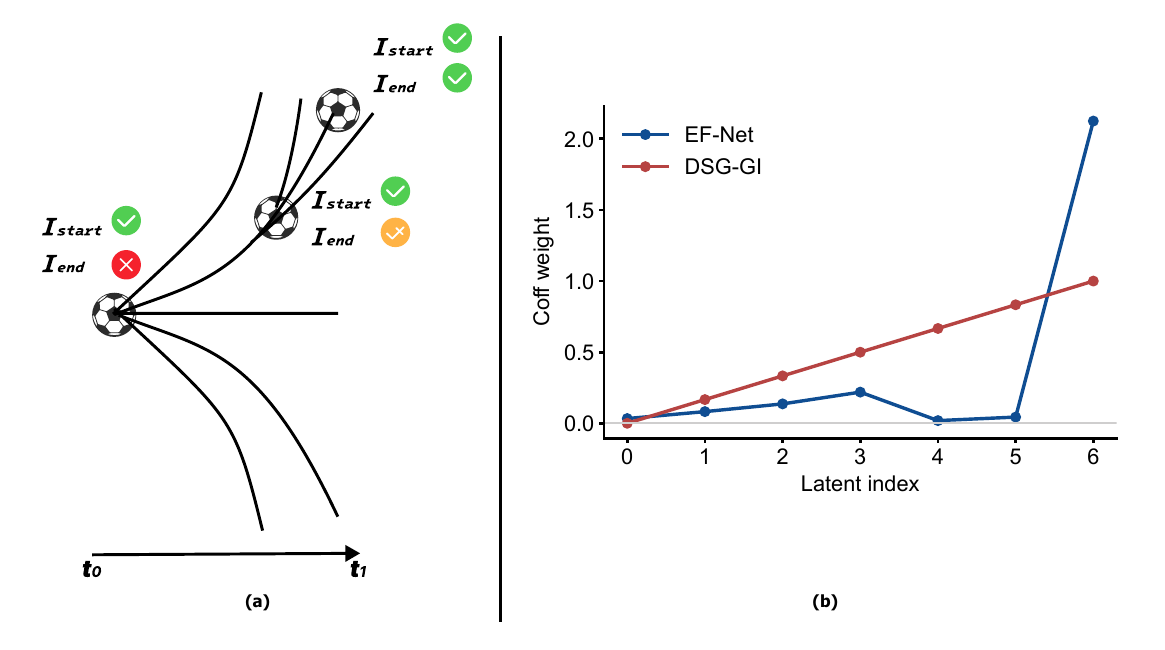} 
\caption{Motivation for temporally progressive target-frame guidance. (a) Early frames should remain anchored to the start frame, while target-frame guidance should progressively increase at later temporal positions to steer the trajectory toward the destination. (b) EF-Net concentrates its temporal weights at the final latent position, whereas DSG-GI progressively increases guidance along the latent temporal axis.}
\label{fig1}
\end{figure}

\begin{figure*}[t]
    \centering
    \includegraphics[width=1.0\textwidth]{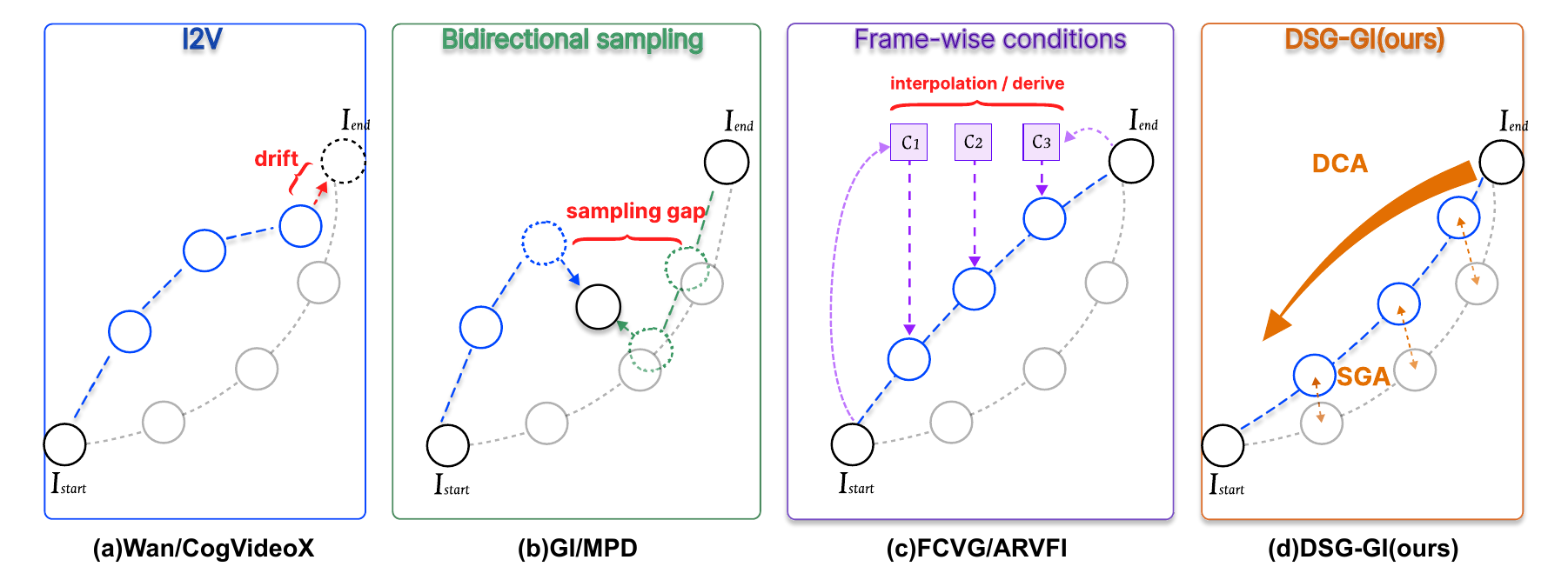}
    \caption{
     (a) Standard I2V generation extrapolates from the start frame and may drift from the target frame. (b) Bidirectional sampling reduces target-frame drift by generating in both temporal directions, but may introduce a discontinuity between the trajectories. (c) Frame-wise methods infer intermediate conditions independently, potentially producing a trajectory that deviates from the ground-truth motion. (d) DSG-GI treats the target frame as the destination of the trajectory: DCA progressively strengthens target-frame guidance, while SGA regularizes the generated trajectory using the semantic-geometric structure of the ground-truth video.
}
    \label{fig2}
\end{figure*}

Large-scale pretrained video diffusion models provide a promising foundation for generative inbetweening~\citep{ho2020denoising,rombach2022high,blattmann2023stable,wan2025wan,yang2025cogvideox}. These models learn rich generative priors from large video corpora and synthesize plausible intermediate content under substantial motion. Recent methods adapt image-to-video diffusion models through start- and target-frame conditioning, bidirectional sampling, or auxiliary frame-wise controls~\citep{danier2024ldmvfi,jain2024video,wang2025generative,zhu2025generative,yang2025vibidsampler,chen2025efvienhancingendframeinjection,ma2026realtime,Ma_2026_CVPR}. Nevertheless, a structural mismatch remains between open-ended video generation and interpolation constrained by a target frame. Image-to-video models learn to generate a plausible future from an initial frame, whereas generative inbetweening requires the sequence to reach a specified target frame. A generated sequence may therefore remain consistent with the start frame while gradually drifting from the target. Figure~\ref{fig2} illustrates this mismatch across representative generative inbetweening settings.

This mismatch presents two coupled challenges. The first is destination-conditioned adaptation. The target frame defines the destination of the sequence and should influence temporal positions with different strengths. As illustrated in Figure~\ref{fig1}(a), early frames should remain strongly anchored to the start-frame context, while target-frame guidance should progressively increase toward later positions to steer the trajectory toward the destination.
The second challenge is semantic-geometric consistency. As illustrated in Figure~\ref{fig2}(b)--(c), individually plausible intermediate frames may form a sequence with discontinuities between sampled trajectories or motion that deviates from the underlying trajectory. Generative inbetweening therefore requires realistic frames and coherent spatio-temporal relations across the sequence.

The work most closely related to ours is EF-VI~\citep{chen2025efvienhancingendframeinjection}, which strengthens destination conditioning through EF-Net. EF-Net estimates correlation weights between the terminal latent and preceding latent positions to determine the contribution of target-frame information at each position. However, Figure~\ref{fig1}(b) shows that EF-Net assigns most of the weight to the terminal latent and suppresses the weights of most preceding latents toward zero. Constraints derived solely from the terminal latent therefore provide limited destination guidance at earlier temporal positions.

To address these challenges, we propose DSG-GI, a generative inbetweening framework that adapts pretrained latent video diffusion models through two complementary components. Destination-Conditioned Adaptation (DCA) encodes the target frame and expands its features along the latent temporal axis, providing sequence-level conditioning that progressively guides generation toward the destination. Semantic-Geometric Alignment (SGA) aligns the token-wise spatio-temporal relational geometry of backbone features with that of a frozen video foundation model and applies a flow-matching-aware latent flow constraint to samples near the data endpoint of the probability path. Together, DCA and SGA encourage the sequence to satisfy the target-frame constraint while preserving coherent semantic-geometric motion structure, without changing the architecture of the pretrained backbone.

Our contributions are as follows. We propose DSG-GI, a framework that adapts pretrained video diffusion models to generation constrained by a target frame, bridging open-ended generation and coherent inbetweening. We introduce two complementary components: DCA provides temporally progressive target-frame guidance, while SGA transfers relation-level video semantics and flow-matching-aware local geometric constraints from frozen auxiliary models to stabilize intermediate motion. Extensive experiments and ablation studies demonstrate improvements in target-frame faithfulness, temporal stability, and perceptual quality on standard and large-motion inbetweening benchmarks.
\section{Related Work}

\paragraph{Video Frame Interpolation.} 
Video frame interpolation has been widely studied in classical and generative settings. Traditional methods synthesize intermediate frames by estimating optical flow, motion kernels, or local correspondences~\citep{niklaus2017video,bao2019depth,jiang2018super,niklaus2020softmax,sim2021xvfi,reda2022film,kong2022ifrnet,li2023amt,hu2024iq,liu2024sparse}. These methods generally perform well under small or moderate motion, but may struggle with large displacements, occlusions, non-rigid deformations, and complex scene changes that make low-level correspondence ambiguous. Recent generative inbetweening methods use pretrained image-to-video diffusion models conditioned on the start and target frames, sometimes combined with bidirectional sampling or auxiliary controls~\citep{voleti2022mcvd,danier2024ldmvfi,jain2024video,feng2024explorative,wang2025generative,zhu2025generative,chen2025efvienhancingendframeinjection,ma2026realtime,Ma_2026_CVPR}. Despite producing visually plausible frames, these methods still face challenges in maintaining temporal consistency and satisfying the target-frame constraint. These limitations motivate the destination-conditioned adaptation and semantic-geometric alignment in DSG-GI.

\paragraph{Controllable Video Generation.} 
Controllable video generation methods incorporate depth, pose, segmentation, trajectories, or optical flow into pretrained diffusion models~\citep{zhang2023adding,mou2024t2i,guo2023animatediff,he2024cameractrl,wang2024motionctrl,zhang2024controlvideo,zhang2025tora,lin2025ctrl,peng2024controlnext}. Representation alignment similarly transfers knowledge from pretrained visual encoders by aligning intermediate features~\citep{zhang2026videorepa}. These strategies improve motion fidelity and structural consistency, but may require dense frame-wise supervision or additional data. Building on these ideas, DSG-GI treats the target frame as a sequence-level destination condition and combines relation-level representation alignment with flow-matching-aware local geometric consistency. This design regularizes intermediate motion without requiring dense control annotations during inference.
\begin{figure*}[t]
\centering
\includegraphics[width=1.0\textwidth]{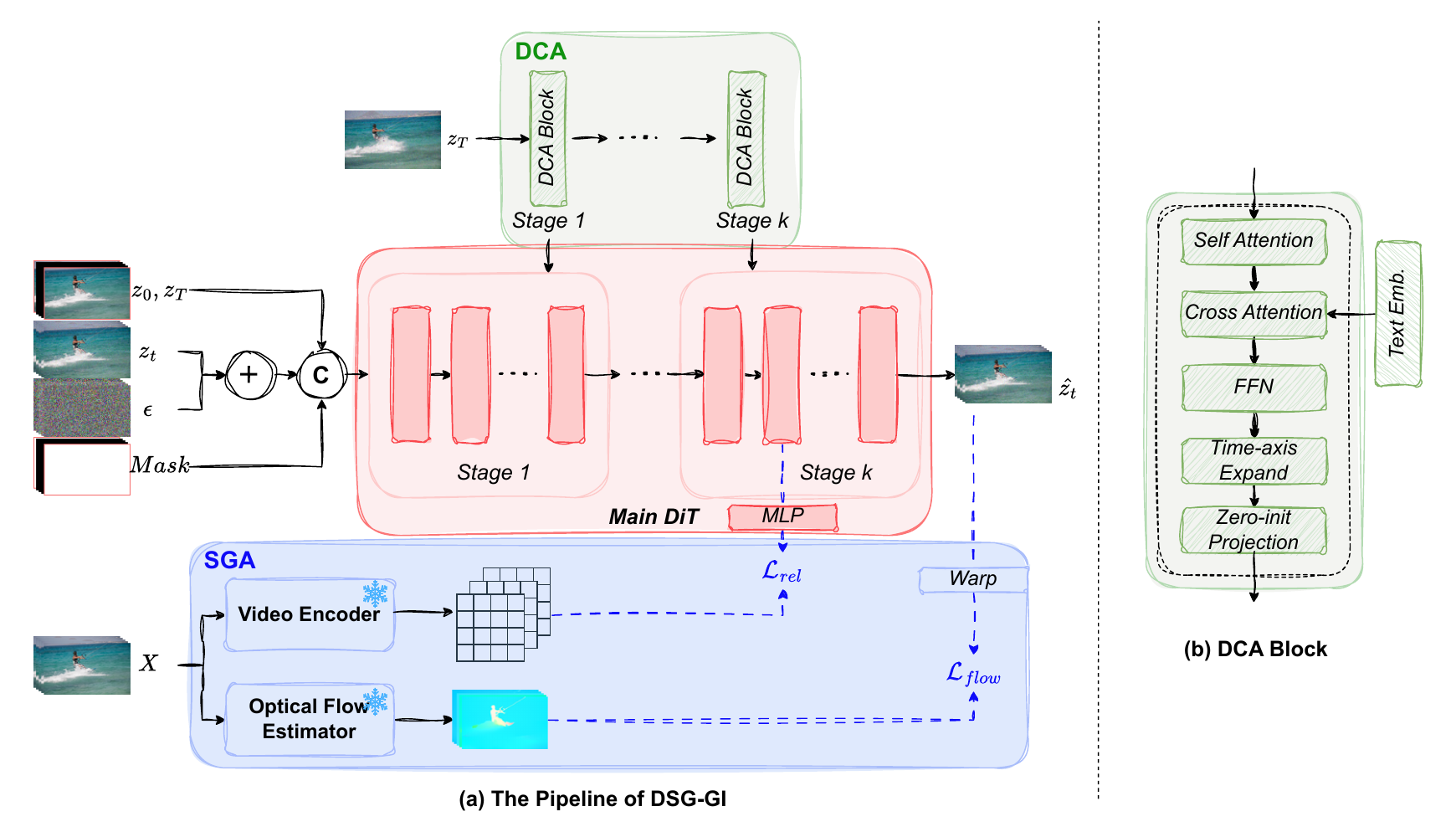} 
\caption{
Overview of DSG-GI. (a) DCA injects temporally progressive target-frame guidance into multiple DiT stages, while SGA regularizes intermediate generation through relation-level semantic alignment and flow-matching-aware latent flow consistency. (b) The DCA block encodes the target-frame condition, expands it along the temporal axis, and applies a zero-initialized projection.
}
\label{fig3}
\end{figure*}

\section{Method}

\subsection{Preliminaries: Flow-Matching Video Diffusion}

DSG-GI builds on a pretrained latent image-to-video transformer optimized with a conditional flow-matching objective~\citep{liu2022flow,rombach2022high,peebles2023scalable}. Given a training video $\bm{X}=\{I_0,I_{1/T},\ldots,I_1\}$, a video VAE encoder $\mathcal{E}$ maps the pixel-space sequence to the data latent $\bm{z}_0=\mathcal{E}(\bm{X})$. Flow matching defines a continuous probability path between $\bm{z}_0$ and a Gaussian reference sample $\bm{\epsilon}\sim\mathcal{N}(0,\bm{I})$. At a sampled path time $t$, the latent state is
\begin{equation}
\bm{z}_t
=
\alpha_t\bm{z}_0
+
\sigma_t\bm{\epsilon},
\end{equation}
where $\alpha_t$ and $\sigma_t$ specify the interpolation schedule along the probability path.

The flow predictor $\mathcal{D}_\theta$ takes the path state $\bm{z}_t$, the flow-matching time $t$, and the conditioning inputs $(I_0,p)$, where $p$ denotes the optional text condition of the pretrained image-to-video backbone. It predicts a vector field with target $\bm{y}_t$, defined by the path parameterization of the pretrained backbone. Training minimizes the mean squared error between the predicted and target vector fields:
\begin{equation}
\mathcal{L}_{\mathrm{FM}}
=
\mathbb{E}_{\bm{X},t,\bm{\epsilon}}
\left[
\left\|
\mathcal{D}_\theta
\left(
\bm{z}_t,t,I_0,p
\right)
-
\bm{y}_t
\right\|_2^2
\right].
\end{equation}

\subsection{Destination-Conditioned Adaptation}
Inspired by EF-VI~\citep{chen2025efvienhancingendframeinjection}, we design Destination-Conditioned Adaptation (DCA) as a lightweight mechanism for injecting target-frame features. The pretrained backbone integrates these features, while SGA regularizes the learned representations. This separation preserves the representation and integration capabilities of the backbone without adding a feature-fusion module. AlignVid~\citep{liualignvid} further shows that pretrained models exhibit coarse foreground--background separation, with most foreground-sensitive blocks concentrated in the earlier half of the network. This distribution motivates injecting target-frame features across multiple stages instead of a single fixed layer. DCA therefore converts the target frame $I_1$ into a sequence-shaped control signal and injects it into different stages of the DiT backbone.

Let the pretrained DiT backbone contain $N$ transformer blocks~\citep{vaswani2017attention,peebles2023scalable}. DCA contains $K$ lightweight target-frame control blocks. To distribute target-frame guidance across the backbone, we divide it into $K$ consecutive stages:
\begin{equation}
\mathcal{S}_k=\left\{b\mid \left\lfloor \frac{kN}{K}\right\rfloor \leq b < \left\lfloor \frac{(k+1)N}{K}\right\rfloor\right\},\quad k=0,\ldots,K-1.
\end{equation}

Each stage receives a target-frame control feature. The target frame is encoded into latent tokens and processed by adapter blocks to produce stage-wise features $\bm{c}_k$. DCA converts this static reference into sequence-level conditioning by expanding each feature along the latent temporal axis with a position-dependent weight:
\begin{equation}
\tilde{\bm{c}}_{k,\tau}=w(\tau)\bm{c}_k,\quad w(\tau)\in\{0,1/T,\ldots,1\}.
\end{equation}

The temporally expanded feature $\tilde{\bm{c}}_{k}$ is projected by a zero-initialized linear layer and injected into the corresponding DiT stage:
\begin{equation}
\bm{h}_b \leftarrow \bm{h}_b + \alpha \mathcal{P}_k(\tilde{\bm{c}}_{k}), \quad b\in\mathcal{S}_k,
\end{equation}
where $\bm{h}_b$ is the hidden representation of the $b$-th DiT block, $\mathcal{P}_k$ is the zero-initialized projection, and $\alpha$ controls the injection strength.


\subsection{Semantic-Geometric Alignment}
DCA improves target-frame reachability but does not uniquely determine the intermediate content: multiple plausible trajectories may connect the same endpoints. Existing I2V models rely primarily on low-level appearance objectives, which encourage realistic individual frames without explicitly preserving sequence-level semantic relations or motion structure. Semantic-Geometric Alignment (SGA) addresses this limitation by aligning the spatio-temporal relational geometry of backbone representations with that of a frozen video foundation encoder. It also provides local geometric supervision through a latent flow constraint when the estimated data latent is sufficiently reliable.

\paragraph{Relation-Level Semantic Geometry.}
Inspired by VideoREPA~\citep{zhang2026videorepa}, we use intermediate DiT features as student representations for semantic alignment. However, the TRD loss in VideoREPA targets higher-level constraints, such as physical regularities of motion in T2V generation, and may be less suited to generative inbetweening. Here, the start and target frames constrain the semantic evolution of the intermediate sequence, so the generative model and video understanding model are expected to capture similar semantic changes over time. We therefore use interpolation instead of convolution to spatially downsample the intermediate DiT features to the teacher feature resolution and apply a margin-relaxed L1 loss to their global relation matrices to preserve the spatio-temporal relational structure.

Let $\mathcal{F}_\psi(\cdot)$ denote a frozen video foundation encoder, such as VideoMAEv2~\citep{wang2023videomae} or V-JEPA2~\citep{assran2025v}, and let $\bm{A}_\theta$ denote intermediate DiT features. We transform $\bm{A}_\theta$ with a lightweight MLP and interpolate the feature volume to the spatio-temporal resolution of the teacher features. For simplicity, $\bm{A}_\theta$ also denotes the transformed features below.

We extract teacher features from the resized ground-truth video:
\begin{equation}
\bm{U}
=
\mathcal{F}_\psi
\left(
\mathrm{Resize}(\bm{X})
\right).
\end{equation}

We reshape the processed student and teacher features into spatio-temporal token sets. Let $\bm{S}$ and $\bm{T}$ denote the normalized student and teacher token matrices, respectively:
\begin{equation}
\bm{S}
=
\mathrm{Norm}_2
\left(
\mathrm{Flat}(\bm{A}_\theta)
\right),
\qquad
\bm{T}
=
\mathrm{Norm}_2
\left(
\mathrm{Flat}(\bm{U})
\right).
\end{equation}

We compute token-wise Gram matrices to capture global pairwise relations across space and time:
\begin{equation}
\bm{R}^S
=
\bm{S}\bm{S}^\top,
\qquad
\bm{R}^T
=
\bm{T}\bm{T}^\top.
\end{equation}

The semantic-geometric relation loss encourages the student to preserve the global relational structure of the teacher through a margin-relaxed objective:
\begin{equation}
\mathcal{L}_{\mathrm{rel}}
=
\frac{1}{B M^2}
\sum_{b,i,j}
\max
\left(
\left|
\bm{R}^S_{b,ij}
-
\bm{R}^T_{b,ij}
\right|
-
m,
0
\right),
\end{equation}
where $M$ is the number of spatio-temporal tokens and $m$ is a scalar tolerance margin. The normalization averages the loss over the batch and all token pairs.

\paragraph{Flow-Matching-Aware Local Geometry.}
The relation loss captures global spatio-temporal semantics, while local motion consistency provides complementary geometric supervision near the data endpoint of the flow-matching path. Let $\sigma_t$ denote the coefficient of the Gaussian reference sample at the sampled flow-matching time $t$. We activate the flow regularizer only when $\sigma_t$ is below a predefined threshold $\gamma$, restricting geometric supervision to states where the estimated data latent is sufficiently reliable. For each active sample, the scheduler converts the predicted vector field into an estimate of the data latent $\hat{\bm{z}}_0$.

A frozen optical-flow estimator computes forward and backward flows between adjacent ground-truth frames. We compose these flows according to the latent temporal stride and resize them to the latent spatial resolution, yielding $\bm{f}^{+}_{\tau}$ and $\bm{f}^{-}_{\tau}$. The forward--backward consistency mask is
\begin{equation}
\bm{M}_{\tau}
=
\mathbb{I}
\left[
\sqrt{
\left\|
\bm{f}^{+}_{\tau}
+
\mathcal{W}
\left(
\bm{f}^{-}_{\tau},
\bm{f}^{+}_{\tau}
\right)
\right\|_{2}^{2}
+
\varepsilon
}
<
\delta
\right],
\end{equation}
where the backward flow is warped by the forward flow and $\delta$ is the consistency threshold. The mask retains locations with consistent forward and backward flows. For each adjacent pair of latent frames, we compute the residual between the current frame and the next frame warped by the forward flow:
\begin{equation}
\bm{r}_{\tau}
=
\hat{\bm{z}}_{0,\tau}
-
\mathcal{W}
\left(
\hat{\bm{z}}_{0,\tau+1},
\bm{f}^{+}_{\tau}
\right).
\end{equation}
Let $N_{\mathrm{pair}}$ denote the number of adjacent latent-frame pairs. The flow loss is defined as
\begin{equation}
\mathcal{L}_{\mathrm{flow}}
=
\mathbb{I}\left[\sigma_t < \gamma\right]
\frac{1}{N_{\mathrm{pair}}}
\sum_{\tau=0}^{N_{\mathrm{pair}}-1}
\mathrm{mean}
\left[
\bm{M}_{\tau}
\odot
\sqrt{\bm{r}_{\tau}^{2}+\varepsilon}
\right].
\end{equation}
The mask $\bm{M}_{\tau}$ suppresses regions with inconsistent forward and backward flows, and the masked residuals are averaged over adjacent latent-frame pairs. Here, $t$ indexes the sampled flow-matching time, $\tau$ indexes the latent-frame pairs, and $\mathcal{W}$ denotes bilinear warping. The loss is active only when $\sigma_t<\gamma$ and is zero otherwise.

The final SGA term is
\begin{equation}
\mathcal{L}_{\mathrm{SGA}}=\beta_{\mathrm{rel}}\mathcal{L}_{\mathrm{rel}}+\beta_{\mathrm{flow}}\mathcal{L}_{\mathrm{flow}},
\end{equation}
with $\beta_{\mathrm{rel}}$ and $\beta_{\mathrm{flow}}$ controlling the relative weights of the relation-level and flow-aware components, respectively.

The complete training objective is
\begin{equation}
\mathcal{L}=\mathcal{L}_{\mathrm{FM}}+\mathcal{L}_{\mathrm{SGA}}.
\end{equation}

DCA and SGA serve complementary roles in DSG-GI. DCA turns the target frame into a temporally progressive destination prior that steers the generated sequence toward the target. SGA regularizes the intermediate transition by aligning global spatio-temporal relations and enforcing local motion consistency. Together, they improve target-frame faithfulness and semantic-geometric coherence. The frozen video encoder and optical-flow estimator used by SGA are required only during training and add no inference-time cost.

\section{Experiments}

\begin{table*}[t]
\centering
\resizebox{\textwidth}{!}{
\begin{tabular}{lcccc|cccc}
\toprule
\multirow{2}{*}{Methods}
& \multicolumn{4}{c|}{FCVG}
& \multicolumn{4}{c}{TGI-Bench} \\
\cmidrule(lr){2-5} \cmidrule(lr){6-9}
& LPIPS $\downarrow$
& FID $\downarrow$
& FVD $\downarrow$
& FVMD $\downarrow$
& LPIPS $\downarrow$
& FID $\downarrow$
& FVD $\downarrow$
& FVMD $\downarrow$ \\
\midrule
FILM (ECCV'22)
& \underline{0.1540} & 25.02 & 472.8 & 8254.0
& \textbf{0.2293} & 26.39 & 568.3 & 15714.3 \\
AMT (CVPR'23)
& 0.2050 & 43.18 & 529.9 & 8174.6
& 0.3006 & 41.28 & 669.2 & 14857.5 \\
SPEED (ACM MM'26)
& \textbf{0.1501} & 25.00 & 488.0 & 5074.5
& \underline{0.2371} & 26.99 & 587.0 & 9807.6 \\
\midrule
GI (ICLR'25)
& 0.2181 & 30.68 & 498.4 & 5542.7
& 0.2979 & 24.83 & 414.8 & 8637.6 \\
FCVG (CVPR'25)
& 0.1831 & 24.09 & 392.0 & 5586.5
& 0.2799 & 22.22 & 365.1 & 8386.5 \\
MPD (ICLR'26)
& 0.2834 & 27.25 & 449.0 & 6493.1
& 0.3371 & 23.64 & 395.2 & \textbf{7141.2} \\
EF-VI (arXiv'25)
& 0.1972 & \underline{22.95} & \underline{339.4} & \underline{4627.3}
& 0.2554 & \underline{19.48} & \underline{331.0} & 7948.8 \\
\midrule
DSG-GI
& 0.1595 & \textbf{20.84} & \textbf{330.7} & \textbf{4437.8}
& 0.2414 & \textbf{18.61} & \textbf{291.7} & \underline{7586.6} \\
\bottomrule
\end{tabular}
}
\caption{Quantitative comparison on FCVG and TGI-Bench. The best and second-best results for each metric are highlighted in \textbf{bold} and \underline{underline}, respectively. $\downarrow$ indicates lower is better.}
\label{tab:main_results}
\end{table*}

\begin{figure*}[t]
\centering
\includegraphics[width=1.0\textwidth]{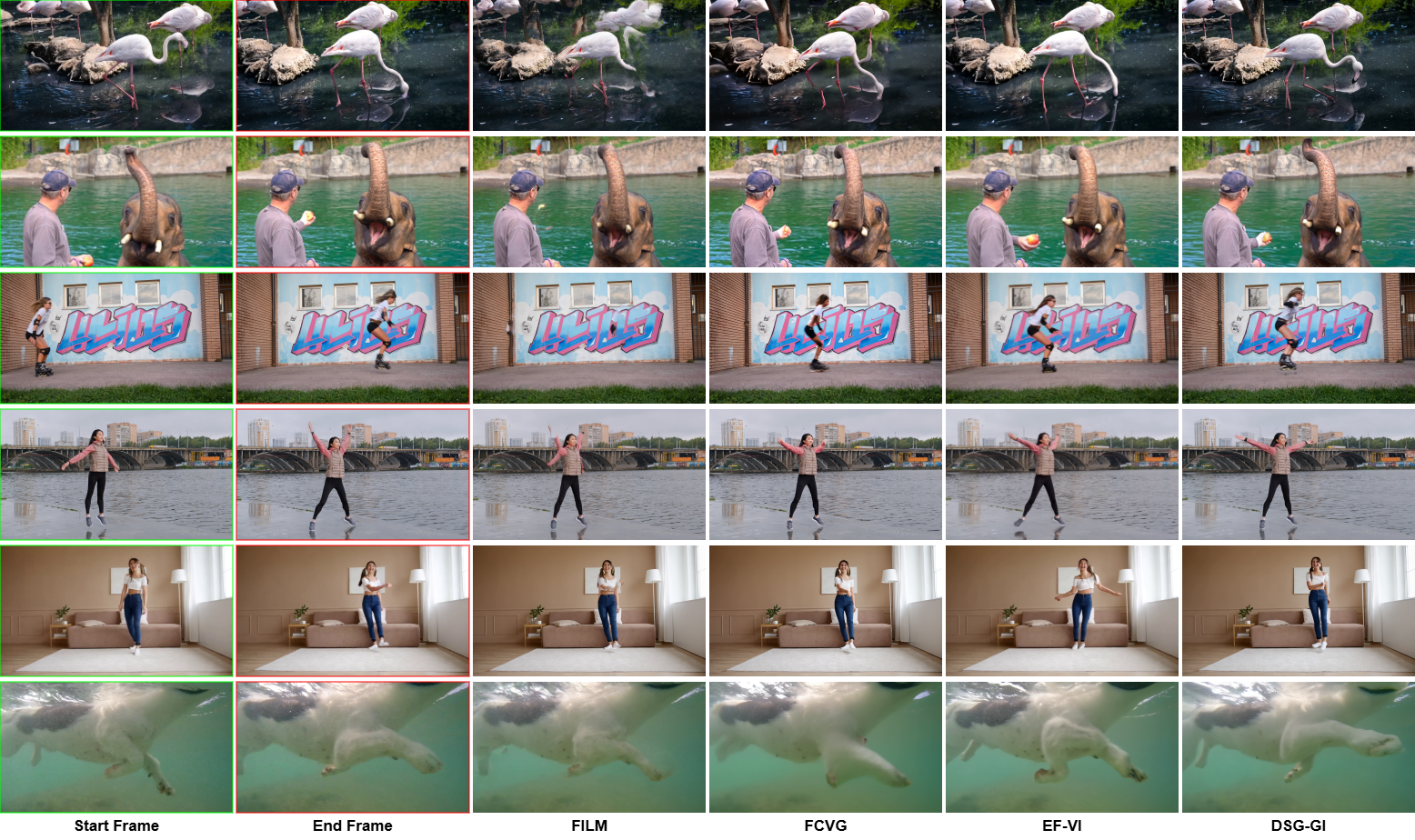} 
\caption{Qualitative comparison with state-of-the-art video interpolation and generative inbetweening methods.}
\label{qualitative_results}
\end{figure*}

\begin{table}[t]
\centering
\resizebox{\columnwidth}{!}{
\begin{tabular}{lcccc}
\toprule
Method
& \begin{tabular}{c}Base Model\\Parameters\end{tabular}
& Video Size
& Steps
& Time (s) \\
\midrule
GI     & 1.5B & 49$\times$1024$\times$576 & 50 & 1707.38 \\
FCVG   & 1.5B & 49$\times$1024$\times$576 & 25 & 364.50 \\
MPD    & 1.5B & 49$\times$1024$\times$576 & 50 & 1037.24 \\
EF-VI  & 5B   & 49$\times$720$\times$480  & 50 & 400.73 \\
DSG-GI & 1.3B & 49$\times$1024$\times$576 & 50 & 180.10 \\
\bottomrule
\end{tabular}
}
\caption{Comparison of base model parameters and inference time across video inbetweening methods.}
\label{tab:inference_time}
\end{table}

\subsection{Experimental Setting}

\paragraph{Datasets.}
For training, we collect 20k real-world video clips from Videvo\footnote{\url{https://www.videvo.net/}}, a publicly available stock-media platform. The clips cover human actions, animal motion, vehicle motion, and natural phenomena, providing diverse motion patterns for adapting the pretrained video generation model. We resize each clip to $1024\times576$ and use its first 25 frames for training. Qwen3-VL~\citep{bai2025qwen3} generates a text description from the start and target frames of each training clip, which serves as the text condition. For evaluation, we follow recent generative inbetweening studies and use the TGI-Bench and FCVG test sets~\citep{zhu2025generative,choi2026anchoring}. These benchmarks cover standard interpolation and large-motion generative inbetweening.

\paragraph{Baselines.}
We compare DSG-GI with seven VFI methods. The three non-generative methods are FILM~\citep{reda2022film}, AMT~\citep{li2023amt}, and SPEED~\citep{zhang2026speed}; the four generative methods are GI~\citep{wang2025generative}, FCVG~\citep{zhu2025generative}, MPD~\citep{jeon2026motion}, and EF-VI~\citep{chen2025efvienhancingendframeinjection}. We evaluate all baselines using their official default settings and recommended configurations.

\paragraph{Metrics.}
We evaluate DSG-GI with four complementary metrics: LPIPS, FID, FVD, and FVMD~\citep{zhang2018unreasonable,heusel2017gans,unterthiner2019fvd,liu2024fr}. LPIPS measures perceptual similarity between generated and ground-truth frames. FID compares feature distributions of generated and real frames to assess visual quality. FVD assesses video quality through differences in spatio-temporal feature distributions, while FVMD evaluates motion quality through differences in motion feature distributions.

\paragraph{Implementation Details.}
DSG-GI is initialized from Wan2.1-Inp-1.3B~\citep{wan2025wan}, a pretrained latent video diffusion transformer based on flow matching. DCA uses $K=2$ target-frame control stages with an injection scale of $\alpha=1.0$. For $\mathcal{L}_{\mathrm{rel}}$, we use intermediate features from the 17th DiT block, VideoMAEv2 as the frozen teacher, and a relation margin of $m=0.1$. For $\mathcal{L}_{\mathrm{flow}}$, we use RAFT~\citep{teed2020raft} as the frozen optical-flow estimator, a latent temporal stride of $4$, a forward--backward consistency threshold of $\delta=1.0$, a reference-sample coefficient threshold of $\gamma=0.2$ along the flow-matching path, and $\varepsilon=10^{-6}$. We train on a single NVIDIA RTX 5880 Ada GPU for 5k iterations with a total batch size of $4$. We use AdamW, a variant of Adam~\citep{kingma2014adam}, with a learning rate of $1\times10^{-5}$ and weight decay of $1\times10^{-2}$. The loss weights are $\beta_{\mathrm{rel}}=5\times10^{-2}$ and $\beta_{\mathrm{flow}}=1\times10^{-4}$. Inference uses $50$ sampling steps.

\subsection{Comparison with the State of the Art}
\paragraph{Quantitative Comparison.}
Table~\ref{tab:main_results} compares DSG-GI with non-generative and generative inbetweening methods on FCVG and TGI-Bench. DSG-GI achieves the lowest FID and FVD on both datasets, the lowest FVMD on FCVG, and the second-lowest FVMD on TGI-Bench. Among generative methods, it also achieves the lowest LPIPS on both datasets.

Among non-generative methods, SPEED achieves the lowest LPIPS on FCVG and FILM on TGI-Bench. Both obtain lower LPIPS than DSG-GI, but substantially higher FID and FVD. AMT is less competitive across all four metrics. These results distinguish frame-level perceptual similarity from distribution-level video quality: lower LPIPS does not necessarily correspond to smaller differences in spatio-temporal or motion feature distributions.

DSG-GI outperforms GI, FCVG, and EF-VI on all reported metrics across both datasets. Relative to EF-VI, it reduces FVD from $339.4$ to $330.7$ on FCVG and from $331.0$ to $291.7$ on TGI-Bench. MPD achieves the lowest FVMD on TGI-Bench, while DSG-GI obtains substantially lower LPIPS, FID, and FVD. These results support the effectiveness of DSG-GI in balancing frame-level perceptual similarity, video quality, and motion quality.

\paragraph{Qualitative Comparison.}
Figure~\ref{qualitative_results} compares challenging examples involving large human motion, sports actions, indoor scenes, and non-rigid animal motion. FILM preserves local appearance in simpler regions, but low-level correspondence errors under large motion gaps often lead to ghosting, blurred structures, or inaccurate poses. FCVG generates more plausible content than warping-based interpolation, yet its frame-wise conditions may produce intermediate frames that deviate from the desired motion path or contain unstable object geometry. EF-VI strengthens the target-frame constraint, but complex motion can still cause temporal discontinuities and pose distortions. DSG-GI produces more coherent transitions between the start and target frames in these examples. \textbf{Additional video comparisons are provided in the supplementary material.}

\paragraph{Computational Efficiency.}
Table~\ref{tab:inference_time} reports model size and inference time. We evaluate all I2V-DM-based methods on a single NVIDIA RTX 5880 Ada GPU using their official sampling-step settings. Despite its relatively small backbone, GI has the longest measured inference time. Bidirectional sampling requires forward and backward model evaluations, and GI repeats the sampling update five times per step to improve prediction fusion~\citep{wang2025generative}. These operations increase computational cost beyond what backbone size alone suggests. DSG-GI uses the smallest backbone, with $1.3$B parameters, and takes $180.10$ s to generate a 49-frame video at $1024\times576$. It has the shortest measured inference time under the reported configurations, although FCVG uses fewer sampling steps and EF-VI generates videos at a lower resolution.

\subsection{Ablation Study}


\begin{table}[t]
\centering
\resizebox{\columnwidth}{!}{
\begin{tabular}{ccc|cccc}
\toprule
\multirow{2}{*}{DCA} 
& \multicolumn{2}{c|}{SGA} 
& \multirow{2}{*}{LPIPS $\downarrow$}
& \multirow{2}{*}{FID $\downarrow$}
& \multirow{2}{*}{FVD $\downarrow$}
& \multirow{2}{*}{FVMD $\downarrow$} \\
\cmidrule(lr){2-3}
& $\mathcal{L}_{\mathrm{rel}}$ 
& $\mathcal{L}_{\mathrm{flow}}$ 
&  &  &  &  \\
\midrule
 &  &  & 0.1706 & 21.90 & 335.3 & 4572.7 \\
\checkmark &  &  & 0.1726 & 22.19 & 327.6 & 4930.7 \\
\checkmark & \checkmark &  & 0.1635 & \underline{21.15} & \textbf{324.4} & \underline{4551.0} \\
\checkmark &  & \checkmark & \underline{0.1623} & 21.17 & \underline{326.2} & 4572.6 \\
\checkmark & \checkmark & \checkmark & \textbf{0.1595} & \textbf{20.84} & 330.7 & \textbf{4437.8} \\
\bottomrule
\end{tabular}
}
\caption{Quantitative ablation of DCA and the SGA losses on the FCVG test set.}
\label{tab:ablation}
\end{table}

\begin{figure}[t]
\centering
\includegraphics[width=0.45\textwidth]{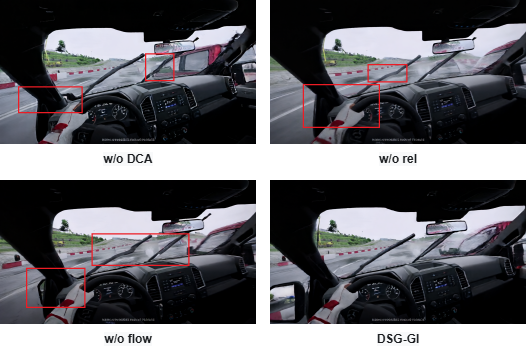} 
\caption{Qualitative ablation of DCA and the SGA losses.}
\label{Ablation}
\end{figure}

We conduct all ablation experiments on the FCVG test set to evaluate the contributions of DCA and SGA and examine the design of the relation loss $\mathcal{L}_{\mathrm{rel}}$.

\paragraph{Component Ablation.}
Table~\ref{tab:ablation} evaluates DCA and the SGA losses, $\mathcal{L}_{\mathrm{rel}}$ and $\mathcal{L}_{\mathrm{flow}}$, on the FCVG test set. The baseline fine-tunes Wan2.1-Inp-1.3B using only the flow-matching objective. Adding DCA alone worsens three of the four metrics: LPIPS increases from $0.1706$ to $0.1726$, FID from $21.90$ to $22.19$, and FVMD from $4572.7$ to $4930.7$, although FVD decreases from $335.3$ to $327.6$. Stronger destination conditioning alone therefore does not consistently improve perceptual quality or motion coherence.

Adding either SGA loss improves all four metrics relative to DCA alone. The relation loss achieves the lowest FVD of $324.4$, while the flow loss reduces LPIPS to $0.1623$ and FVMD to $4572.6$. Combining both losses yields the lowest LPIPS, FID, and FVMD of $0.1595$, $20.84$, and $4437.8$, respectively. Although the full model has a higher FVD than either single-loss variant, it improves all four metrics over the fine-tuning-only baseline.

These findings suggest that DSG-GI benefits from combining destination guidance with constraints on intermediate evolution. DCA supplies target-frame information, but does not by itself ensure coherent intermediate motion. SGA complements this guidance by preserving global spatio-temporal relations and enforcing local geometric consistency, helping produce perceptually better and more coherent transitions.

Figure~\ref{Ablation} illustrates the qualitative effects of these components. The labels ``w/o DCA'', ``w/o rel'', and ``w/o flow'' denote the removal of DCA, $\mathcal{L}_{\mathrm{rel}}$, and $\mathcal{L}_{\mathrm{flow}}$, respectively; ``DSG-GI'' denotes the full model. Without DCA, object layouts become inconsistent and the transition deviates from the desired trajectory. Removing $\mathcal{L}_{\mathrm{rel}}$ weakens global structural coherence, while removing $\mathcal{L}_{\mathrm{flow}}$ leaves local geometric artifacts and motion discontinuities. The full model better preserves scene layout, object boundaries, and motion direction, illustrating the complementary roles of destination guidance, global semantic alignment, and local geometric supervision.

\paragraph{Ablation of $\mathcal{L}_{\mathrm{rel}}$.}
Table~\ref{tab:alignment_variants} compares the TRD loss in VideoREPA, denoted by $\mathcal{L}_{\mathrm{TRD}}$, with our relation loss $\mathcal{L}_{\mathrm{rel}}$ using V-JEPA2 or VideoMAEv2 as the frozen teacher. The VideoMAEv2 configuration corresponds to DSG-GI and achieves the lowest values across all four metrics: LPIPS, FID, FVD, and FVMD are $0.1595$, $20.84$, $330.7$, and $4437.8$, respectively. Compared with the $\mathcal{L}_{\mathrm{TRD}}$ variant, this configuration reduces FVD from $370.8$ to $330.7$ and FVMD from $6398.7$ to $4437.8$, while improving LPIPS and FID. The $\mathcal{L}_{\mathrm{rel}}$ configuration with V-JEPA2 has lower FVD and FVMD than the TRD variant, but higher LPIPS and FID. It also has higher values than the VideoMAEv2 configuration across all four metrics.

\begin{figure}[t]
\centering
\includegraphics[width=0.45\textwidth]{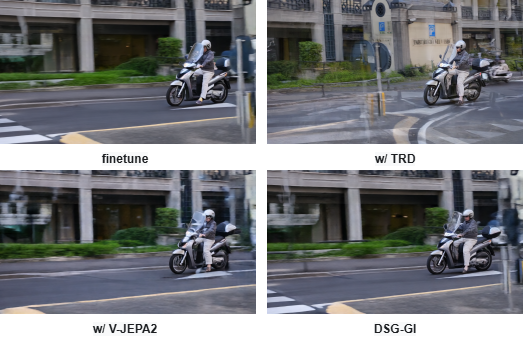} 
\caption{Qualitative comparison of semantic alignment variants.}
\label{fig6}
\end{figure}

Figure~\ref{fig6} compares the fine-tuning baseline and semantic alignment variants.The label ``finetune'' denotes the variant without Relation-Level Semantic Geometry; ``w/ TRD'' uses $\mathcal{L}_{\mathrm{TRD}}$, ``w/ V-JEPA2'' uses $\mathcal{L}_{\mathrm{rel}}$ with V-JEPA2, and ``DSG-GI'' denotes using $\mathcal{L}_{\mathrm{rel}}$ with VideoMAEv2. These results support tailoring semantic alignment to generative inbetweening constrained by endpoint frames. TRD emphasizes higher-level constraints in open-ended T2V generation, whereas the start and target frames restrict the semantic evolution of an inbetweening sequence. Our margin-relaxed relation loss aligns global pairwise relations among spatio-temporal tokens to preserve this structure. Notably, although V-JEPA2 reports stronger performance on video understanding benchmarks with frozen encoders~\citep{assran2025v}, VideoMAEv2 yields better results as a teacher in our experiments. This may reflect differences in their pretraining objectives: VideoMAEv2 reconstructs masked video pixels~\citep{wang2023videomae}, whereas V-JEPA2 predicts masked embeddings in a learned representation space~\citep{assran2025v}. Pixel reconstruction may encourage features that retain appearance and spatial details useful for pairwise spatio-temporal relation alignment.

\begin{table}[t]
\centering
\resizebox{\columnwidth}{!}{
\begin{tabular}{lcccc}
\toprule
Method & LPIPS $\downarrow$ & FID $\downarrow$ & FVD $\downarrow$ & FVMD $\downarrow$ \\
\midrule
$\mathcal{L}_{\mathrm{TRD}}$
& \underline{0.1724} & \underline{22.73} & 370.8 & 6398.7 \\
$\mathcal{L}_{\mathrm{rel}}$ + V-JEPA2
& 0.1786 & 23.16 & \underline{355.5} & \underline{4688.8} \\
$\mathcal{L}_{\mathrm{rel}}$ + VideoMAEv2
& \textbf{0.1595} & \textbf{20.84} & \textbf{330.7} & \textbf{4437.8} \\
\bottomrule
\end{tabular}
}
\caption{Quantitative comparison of semantic alignment variants.}
\label{tab:alignment_variants}
\end{table}

\section{Limitations}
DSG-GI has several limitations. First, SGA depends on a frozen video representation model and a frozen optical-flow estimator during training. Neither is required during inference, but inaccurate teacher features or flow estimates under severe occlusion, rapid non-rigid motion, or non-photorealistic content may weaken the learned constraints. Second, the soft destination prior in DSG-GI does not provide explicit physical reasoning. The model may therefore struggle with highly stochastic events, topological changes, or multiple equally plausible motion paths.

\section{Conclusion}
DSG-GI formulates generative inbetweening as destination-constrained flow matching. DCA provides temporally progressive target-frame guidance, while SGA regularizes intermediate motion through relation-level semantic alignment and flow-matching-aware local geometry. Quantitative and qualitative results demonstrate improvements in perceptual quality, target-frame faithfulness, and motion coherence. These findings support combining destination-conditioned adaptation with semantic-geometric alignment.

{
    \small
    \bibliographystyle{ieeenat_fullname}
    \bibliography{main}
}

\clearpage
\setcounter{page}{1}
\maketitlesupplementary

We provide three supplementary files: this PDF document, \texttt{code.zip} with the training, inference, and evaluation code, and \texttt{video\_demo.zip} with a video demonstration of dynamic comparisons.


\section{Text Conditioning}
\begin{figure}[t]
\centering
\includegraphics[width=0.4\textwidth]{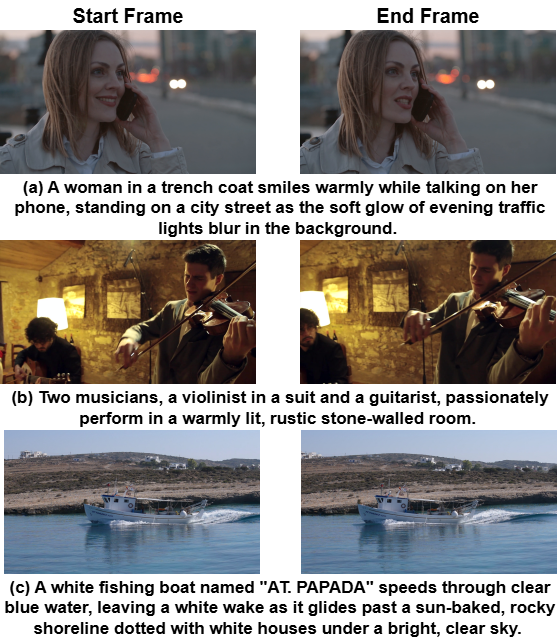} 
\caption{Examples of text prompts generated by Qwen3-VL-8B from start and end frames.}
\label{fig:textcond}
\end{figure}

Following the protocol of EF-VI for using a vision-language model in video inbetweening~\citep{chen2025efvienhancingendframeinjection}, we use Qwen3-VL-8B~\citep{bai2025qwen3} to generate text prompts solely from the start and end frames. We provide each start-end-frame pair together with the instruction: ``This is the first and last frame of a video clip. Describe this video in one short motion caption.'' The resulting caption serves as the text condition during training and inference.
Figure~\ref{fig:textcond} presents examples of the generated text prompts.

\section{More Visual Comparisons}
Figure~\ref{fig:morevis} provides additional intermediate-frame comparisons, and Figure~\ref{fig:multiframes} presents generated sequence examples. The examples cover human actions, animal motion, and moving objects, illustrating how DSG-GI preserves object structure and scene appearance while generating coherent transitions between the start and end frames. The MP4 examples in \texttt{video\_demo.zip} provide more detailed comparisons of motion and temporal consistency.

\section{Failed Cases}
\begin{figure}[t]
\centering
\includegraphics[width=0.5\textwidth]{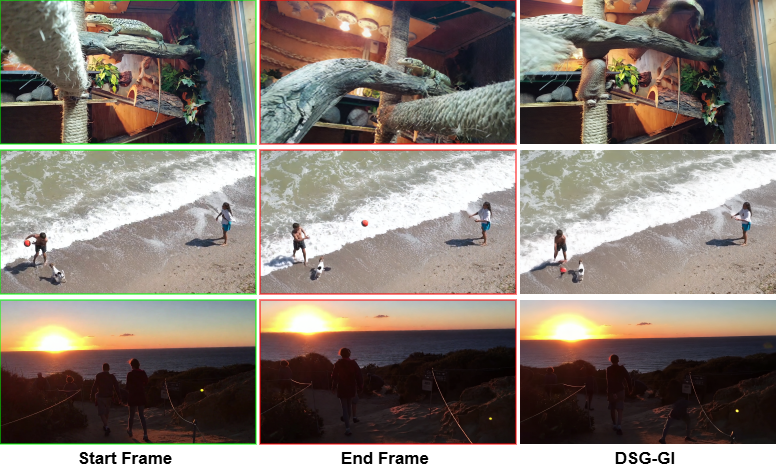} 
\caption{Failure cases of DSG-GI in generating intermediate frames between the start and end frames.}
\label{fig:failed}
\end{figure}

\begin{figure*}[t]
\centering
\includegraphics[width=1.0\textwidth]{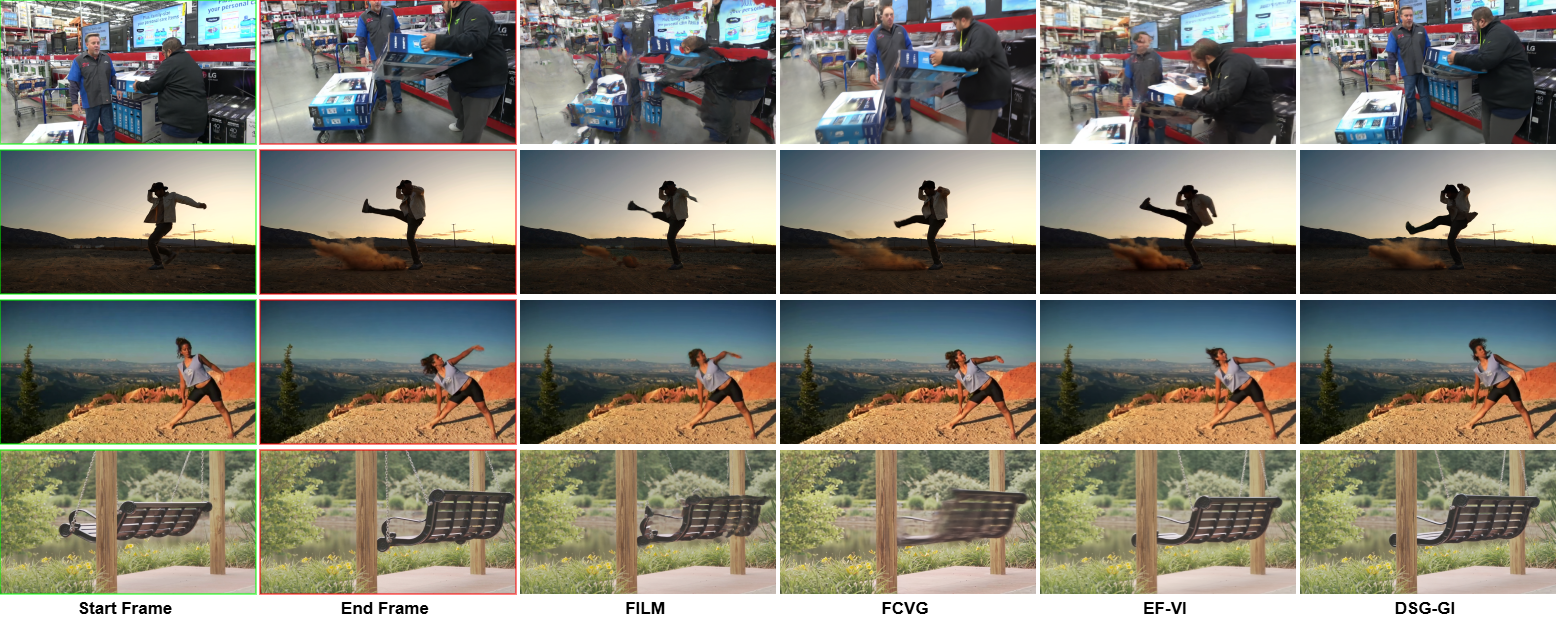} 
\caption{Additional qualitative comparisons with state-of-the-art video interpolation and generative inbetweening methods.}
\label{fig:morevis}
\end{figure*}

\begin{figure*}[t]
\centering
\includegraphics[width=1.0\textwidth]{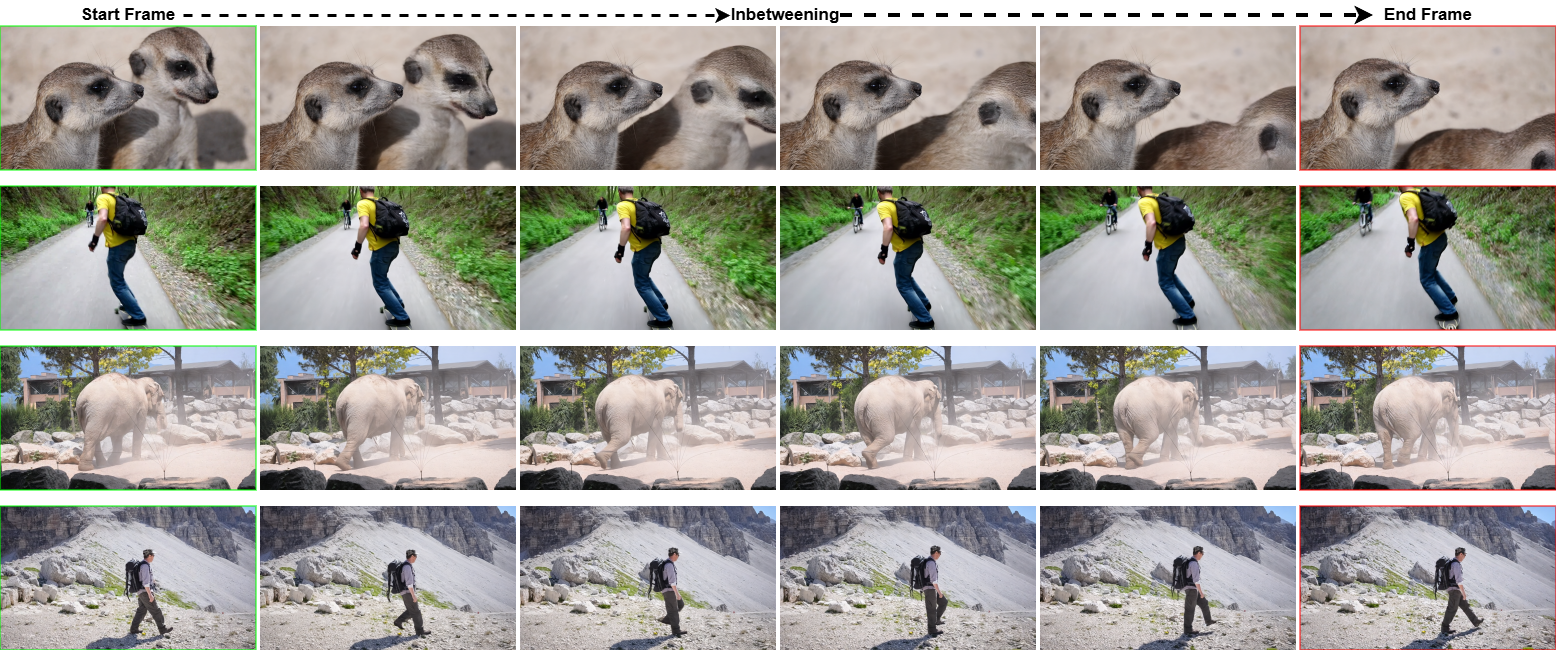} 
\caption{Examples of intermediate frames generated by DSG-GI between the start and end frames.}
\label{fig:multiframes}
\end{figure*}

Figure~\ref{fig:failed} presents failure cases of DSG-GI. In these examples, the model fails to preserve object structure or consistent subject appearance in scenes with substantial viewpoint changes, multiple moving subjects, and complex dynamics. Simultaneous camera and object motion requires coordination of global viewpoint changes with local movement, deformation, and occlusion. Overlapping or changing motion patterns may exceed the learned motion-modeling capabilities of the pretrained backbone.

These failures are consistent with limitations of the adaptation and alignment mechanisms. DCA provides temporally progressive target-frame guidance, but its soft destination prior does not specify intermediate camera or object trajectories. SGA encourages global semantic relations and local flow consistency during training, but does not explicitly disentangle sources of motion. These constraints may therefore be insufficient to resolve ambiguous intermediate dynamics from the endpoints, particularly when camera movement accompanies independently moving objects. Improving the capacity of the backbone to represent coupled motion and explicitly modeling camera and object trajectories are promising directions for addressing these limitations.

\end{document}